# Global Conditioning for Probabilistic Inference in Belief Networks

**Ross D. Shachter**
Department of
Engineering-Economic Systems
Stanford University
Stanford, CA 94305-4025
shachter@camis.stanford.edu

**Stig. K. Andersen**
Dept. of Medical Informatics
and Image Analysis
Aalborg University
DK-9220 Aalborg SO DENMARK
ska@miba.auc.dk

**Peter Szolovits**
Laboratory for Computer Science
Massachusetts Institute of Technology
545 Technology Square
Cambridge, MA 02139
psz@lcs.mit.edu

## Abstract

In this paper we propose a new approach to probabilistic inference on belief networks, global conditioning, which is a simple generalization of Pearl's (1986b) method of loop-cutset conditioning. We show that global conditioning, as well as loop-cutset conditioning, can be thought of as a special case of the method of Lauritzen and Spiegelhalter (1988) as refined by Jensen et al (1990a; 1990b). Nonetheless, this approach provides new opportunities for parallel processing and, in the case of sequential processing, a tradeoff of time for memory. We also show how a hybrid method (Suermondt and others 1990) combining loop-cutset conditioning with Jensen's method can be viewed within our framework. By exploring the relationships between these methods, we develop a unifying framework in which the advantages of each approach can be combined successfully.

## 1 INTRODUCTION

In recent years, there have been intense research efforts to develop efficient methods for probabilistic inference on belief networks. A number of different exact methods have been proposed and implemented to solve this general class of problems. In this paper, we introduce a solution method we call the Clustering Algorithm, a variant of the HUGIN algorithm (Jensen and others 1990a; Jensen and others 1990b), and closely related to a variety of algorithms (Cannings and others 1976; Cannings and others 1978; Lauritzen and others 1990; Shafer and Shenoy 1990; Shenoy 1986). The Clustering Algorithm is not an improvement over the methods listed above, but rather a framework for comparison and integration with the Polytree Algorithm (Kim and Pearl 1983; Pearl 1986b) and Loop-Cutset Conditioning (Pearl 1986a; Peot and Shachter 1991). We show how these techniques are special cases of the Clustering Algorithm. Although the strong relationship between the Clustering and Polytree Algorithms is widely understood, the same has not been true of Loop-Cutset Conditioning, despite our announcements to that effect several years back at the Conference on Uncertainty in Artificial Intelligence.

In the process of demonstrating results about Loop-Cutset Conditioning, we generalize it to an algorithm we call Global Conditioning. Even though Global Conditioning is shown to be a special case of the Clustering Algorithm, the insights it provides suggest new applications with parallel processing and under memory restrictions. We also provide a Clustering Algorithm interpretation to a hybrid algorithm which combines Jensen's method with Loop-Cutset Conditioning (Suermondt and others 1990).

Section 2 defines the notation and terms to be used throughout the paper. Sections 3 and 4 introduce the Clustering Algorithm and the Method of Global Conditioning, respectively, and explore their connection with the Polytree and Loop-Cutset Conditioning Methods. Section 5 presents a parallelizable approach based on Global Conditioning and shows how it can also trade compute time for memory. Section 6 shows dynamic restructuring of cluster trees based on the logic of conditiong, while conclusions and suggestions for future research are presented in Section 7.

## 2 NOTATION AND BASIC FRAMEWORK

In this section, we present our notation and some standard definitions. In particular we introduce the framework for random variables and evidence, and their representation in directed and undirected graphs. We also present an example problem which will be used throughout the paper.

We have a finite set of **elements** $N = \{1, \ldots, n\}$, corresponding to the nodes in a directed acyclic graph. Such a directed graph has many names in the literature, including **belief network** (Pearl 1986b), **prob-** abilistic influence diagram (Shachter 1988), and causal probabilistic network (Jensen and others 1990b). We refer to a particular element with a lower case letter, $j \in N$, and a set of elements using upper case letters, $J \subseteq N$. The parents of a node are denoted by $Pa(j)$, and the parents of a set of nodes, $Pa(J)$ is simply the union of the parents of the nodes in the set. The family of a node $j$ consists of $j$ and its parents, $\{j\} \cup Pa(j)$. An example of a belief network is shown in Figure 1a. It is taken from Lauritzen and Spiegelhalter (1988).

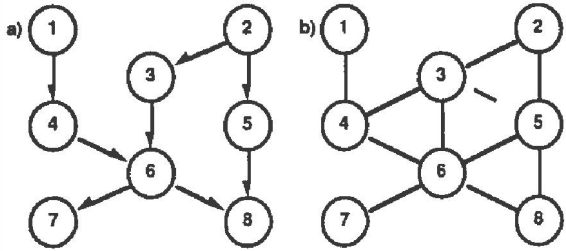

Figure 1: Chest clinic belief network example.

Associated with each element $j$ is a random variable $X_j$ having a finite number of possible values $x_j$. A vector of variables $X_J$ is associated with the set of elements $J$. Each variable $X_j$ has a prior conditional probability distribution, $\Pr\{X_j|X_{Pa(j)}\}$. Let $\epsilon$ be the evidence which has been obtained about the variables $X_N$. Each observation $\epsilon_k$, whether an exact observation or indirect evidence about a variable or set of variables, can be characterized by a likelihood function, $\Pr\{\epsilon_k|X_{J_k}\}$, of the variables $X_{J_k}$.

The elements $N$ can also be represented as the nodes in an undirected graph, called a moral graph. The moral graph is constructed by adding an undirected arc between all nodes with common children, and then replacing the directed arcs with undirected arcs. The moral graph corresponding to the belief network shown in Figure 1a is displayed in Figure 1b, without the dashed (3, 5) arc. Moralizing arcs (3, 4) and (5, 6) were added. A moral graph is said to be chordal or triangulated if there is no cycle of length four or more which does not have an arc between two nonadjacent nodes in the graph. Although the graph shown in Figure 1b is not chordal without the (3, 5) arc, it is chordal with it, or with other arcs which could have been added instead such as (2, 6).

Graphs can also be constructed in which each node corresponds to a set of elements rather than just a single element. A set of elements for this purpose will be called a cluster. An undirected tree of clusters will be called a join tree if every element which appears in more than one cluster appears in every cluster on the path between them (Beeri and others 1983).

A cluster tree for a particular belief network is a join tree in which every family from the belief network is contained in at least one cluster. For example, the four trees drawn in Figure 2 are cluster trees for the belief network drawn in Figure 1a. Cluster trees have been recognized before (as "junction trees") and some of their special properties are well known (Jensen and others 1990a). The moral graph corresponding to a cluster tree consists of nodes for each element with an undirected arc between any two nodes whose corresponding elements appear in the same cluster. Such a moral graph will always be chordal and consistent with the original belief network and its moral graph. The independence properties represented in the cluster tree can also be revealed if, for every pair of neighboring clusters, an intervening cluster containing their intersection is inserted on the arc between them. After the addition of these new clusters, called separation sets, the new graph is still a cluster tree, but now all of the conditional independence is explicitly represented, namely, any two clusters in a cluster tree are conditionally independent given any intervening cluster or the separation set corresponding to any arc in between them. The graph drawn in Figure 2d was constructed from the cluster tree shown in Figure 2a by the addition of the separation sets followed by the deletion of redundant clusters for (3, 6).

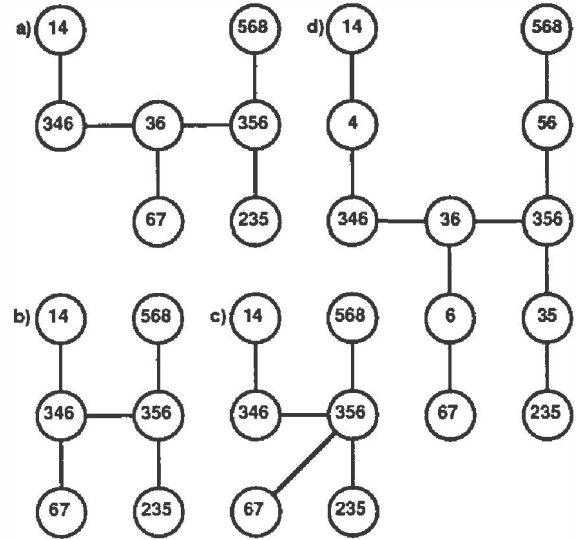

Figure 2: Cluster trees for the chest clinic example.

## 3 THE CLUSTERING ALGORITHM

The Clustering Algorithm performs probabilistic inference by passing messages around cluster trees and propagating the global effects of local information. It is a variation on the algorithm in HUGIN (Jensen and others 1990a; Jensen and others 1990b) but also closely related to other algorithms (Cannings and others 1976; Cannings and others 1978; Lauritzen and others 1990; Shafer and Shenoy 1990; Shenoy 1986). We show that the Polytree Algorithm (Kim and Pearl 1983; Pearl 1986b; Peot and Shachter 1991) can be viewed as a special case of the Clustering Algorithm.

There are only two kinds of data which are truly local in a cluster tree: the conditional probability distributions and likelihood functions for variables in the cluster. Those functions can be set independently for each variable at any time, although there is a global restriction on what constitutes a consistent or coherent set of data for the entire cluster tree, namely $\Pr\{\epsilon\} > 0$. Each variable $X_j$ in the model has a conditional probability distribution $\Pr\{X_j|X_{Pa(j)}\}$ which is assigned to exactly one cluster containing $j$'s family. Likewise for each observation $\epsilon_k$ there is a likelihood function $\Pr\{\epsilon_k|X_{J_k}\}$ showing how the probability for the evidence depends on the variables $X_{J_k}$. Each likelihood function is assigned to exactly one cluster containing $J_k$. The product of the conditional probability distributions and likelihood functions assigned to cluster $S_i$ is called the **potential function** $\Psi_i = \Psi_i(X_{S_i})$ for cluster $S_i$, and it contains all of the local information for that cluster. (The arguments for $\Psi_i$ will be omitted to improve readability whenever they are unambiguous.)

The joint distribution for the model is the global information obtained by multiplying together all of the local information. For example, the posterior joint distribution for all of the variables is just

$$\Pr\{X_N, \epsilon\} = \prod_{j \in N} \Pr\{X_j|X_{Pa(j)}\} \prod_k \Pr\{\epsilon_k|X_{J_k}\}$$
$$= \prod_i \Psi_i.$$

The posterior joint distribution over the variables $X_J$ can then be computed by marginalizing,

$$\Pr\{X_J, \epsilon\} = \sum_{N \setminus J} \Pr\{X_N, \epsilon\} = \sum_{N \setminus J} \prod_i \Psi_i,$$

where $\sum_{N \setminus J}$ signifies a summation over all possible values of all argument variables except $X_J$.

The local information within the cluster tree is propagated throughout the network via local messages in order to compute these global expressions. Consider two adjacent clusters $S_i$ and $S_j$. If $K_{i|j}$ is the set of cluster indices for all clusters on $S_i$'s side of the (i, j) arc and $K_{j|i}$ is the complementary set of clusters on $S_j$'s side of the arc, define the **cluster message** $M_{ij}$ sent from $S_i$ to $S_j$ to be

$$M_{ij} = \sum_{N \setminus S_j} \prod_{k \in K_{i|j}} \Psi_k.$$

Because the cluster tree is a join tree, the only elements in common between $S_j$ and all of the clusters in $K_{i|j}$ must be in $S_i$, so $M_{ij}$ is a function of variables $X_{S_i \cap S_j}$. As a result, we can compute the **separation set probability** $P_{ij}$, the posterior joint distribution for the separation set for two adjacent clusters, as the product of the two cluster messages on the arc connecting them:

$$P_{ij} = \Pr\{X_{S_i \cap S_j}, \epsilon\} = \sum_{N \setminus (S_i \cap S_j)} \prod_k \Psi_k$$
$$= (\sum_{N \setminus S_j} \prod_{k \in K_{i|j}} \Psi_k)(\sum_{N \setminus S_i} \prod_{k \in K_{j|i}} \Psi_k)$$
$$= M_{ij} M_{ji}.$$

Because all of the local data are partitioned into clusters indexed by $K_{i|j}$ and $K_{j|i}$, the two messages not only contain all relevant information, but they can also be computed asynchronously. In practice, the messages can be computed in terms of other messages,

$$M_{ij} = \sum_{N \setminus S_j} \prod_{k \in K_{i|j}} \Psi_k = \sum_{S_i \setminus S_j} \Psi_i \prod_{k \in K_{i-j}} M_{ki},$$

where $K_{i-j}$ is the set of all clusters adjacent to $S_i$, excepting $S_j$.

All of the incoming cluster messages combined with the local data also allow us to compute the **posterior probability distribution** $P_i = \Pr\{X_{S_i}, \epsilon\}$ for any cluster $S_i$,

$$P_i = \sum_{N \setminus S_i} \prod_k \Psi_k = \Psi_i \prod_{k \in K_i} M_{ki},$$

where $K_i$ is the set of all clusters adjacent to $S_i$. From these expressions it is easy to confirm that for any adjacent clusters $S_i$ and $S_j$,

$$\Pr\{X_{S_i \cap S_j}, \epsilon\} = \sum_{S_i \setminus S_j} P_i = M_{ij} M_{ji} = P_{ij}$$
$$= \sum_{S_j \setminus S_i} P_j = M_{ji} M_{ij} = P_{ji}.$$

There are a number of different strategies to control message passing that constitute the **Clustering Algorithm**. First, because a cluster message summarizes all of the local information on one side of an arc, any change to the local information can require updating of the message. Conversely, if there is no change to any of the local information on one side of an arc, then there will be no change to the cluster message. Second, if all that is required is the computation of the posterior distribution for a particular cluster $S_i$, that can be performed with local computations through a **collect operation** (Jensen and others 1990b):
1. $S_i$ sends request messages to all of its neighbors;
2. They in turn send request messages to all of their other neighbors and so forth;
3. Eventually a request message reaches a cluster with only one neighbor. It either computes and sends a new cluster message in reply or indicates that the old message is still valid;
4. Clusters with more than one neighbor either compute and send a new message or, if they have no changes and neither do any of their other neighbors, indicate that the old message is unchanged;
5. Eventually, $S_i$ will receive a message from each of its neighbors and thus it can compute $P_i$.

Third, if all clusters are to be updated, then a **distribute operation** can also be performed with local operations (Jensen and others 1990b): $S_i$ sends updated cluster messages to each of its neighbors, and they in turn send them to their other neighbors and so forth until every cluster has received such a message.

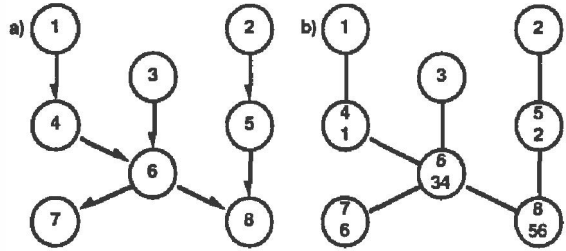

Figure 3: The Polytree Algorithm applied to the chest clinic example.

Finally, initialization simply requires multiplying the factors in each $\Psi_i$ and marking the cluster as changed. If there are no probability factors for a cluster $S_i$, then its $\Psi_i$ should be a scalar 1 and it should be marked as not changed. If there is some evidence to begin with, then it should also be factored into the appropriate $\Psi_i$ and the cluster should be marked as changed.

**The Polytree Algorithm**

The **Polytree Algorithm** for singly-connected belief networks (Kim and Pearl 1983; Pearl 1986b) forms a cluster tree with the same topology as the original network, but with each node $j$ replaced by its family, $S_j = \{j\} \cup Pa(j)$. Because the belief network is singly connected, the resulting cluster tree is a join tree. The cluster conditional probability is simply the node's conditional probabilities, $P_j = \Pr\{X_j | X_{Pa(j)}\}$, and the cluster likelihood function is just the evidence pertaining directly to $X_j$, $L_j = \Pr\{\epsilon_j | X_j\}$. The Clustering Algorithm is then performing precisely the Revised Polytree Algorithm, in which causal support messages are unconditional rather than conditional (Peot and Shachter 1991). Consider the belief network shown in Figure 3a, in which the (2, 3) arc has been eliminated to render the network singly-connected. The cluster tree for the Polytree method is shown in Figure 3b. In general, the following theorem applies.

**Theorem 1 (The Polytree Algorithm)**
*The Polytree Algorithm (Kim and Pearl 1983; Pearl 1986b) in modified form (Peot and Shachter 1991) is a special case of the Clustering Algorithm.*

There are also choices in the representation of the cluster messages. We have presented the messages in **factored form**, which simplifies the revision of prior probability distributions and the retraction of evidence. The messages can also be represented in **joint form** for more efficient calculations. Each cluster $S_i$ remembers its posterior joint distribution $P_i$ and each arc (or separation set) from cluster $S_i$ to cluster $S_j$ remembers the posterior joint distribution last sent, $P_{ij}^{\text{old}}$. The message sent between clusters $S_i$ and $S_j$ is not $M_{ij}$ but rather $P_{ij}^{\text{new}} = M_{ij} M_{ji}$, the separation set joint distribution, which is of the same dimensions. The actual update is just the element-by-element ratio

$$\rho_{ij} = P_{ij}^{\text{new}} \div P_{ij}^{\text{old}},$$

where any value can be used for $0 \div 0$. The posterior distribution for cluster $S_i$ is

$$P_i^{\text{new}} \leftarrow P_i^{\text{old}} \prod_{k \in K_i} \rho_{ij}$$

and

$$M_{ji} M_{ji} \leftarrow \sum_{S_i \setminus S_j} P_i \quad \text{for} \quad j \in K_i,$$

where $K_i$ is the set of all clusters adjacent to $S_i$. There are also tremendous opportunities for savings in the compiled compression of zeros from all of the distributions (Jensen and Andersen 1990), which means that most components of a sparse message and joint distribution are never computed. The only cost comes when evidence is retracted, and the whole network might have to be reinitialized. Initialization is not difficult however, with $P_{ij}^{\text{old}}$ set to scalar 1, and the remaining initialization process similar to the factored form.

## 4  GLOBAL CONDITIONING

The Methods of Global and Loop-Cutset Conditioning are presented in this section. Loop-Cutset Conditioning (Pearl 1986a) is one of the oldest methods for solving multiply-connected belief networks. We show that both conditioning methods can be viewed as special cases of the clustering algorithm.

**The Method of Global Conditioning** provisionally observes a set of variables $X_K$, the **conditioning set**, in order to simplify the belief network, rendering it easier to solve. We will call the original problem the **master problem**, and the simplified subproblems the **instantiated problems**. Although we could conceivably use any technique to solve the instantiated subproblems, we will assume in this paper that they will be solved with the Clustering Algorithm. Although we will not explore methods to select the conditioning set, it is critical to the success of the method and a subject worthy of papers by itself.

Having "observed" $X_K$, we can cut the outgoing arcs from $K$ in the belief network, which is equivalent to separation by $K$ in the moral graph (Shachter 1990a). For example, the multiply-connected graph shown in Figure 1a becomes the singly-connected graph shown in Figure 4a after $X_2$ has been "observed." We can compute the overall desired result by considering all possible cases for $X_K$ and weighting by their probabilities, using the "Law of Total Probability,"

$$\Pr\{X_J, \epsilon\} = \sum_{x_K} \Pr\{X_J, X_K = x_K, \epsilon\}.$$

The main problem with the method is that the number of terms in the sum grows exponentially with the number of elements in the conditioning set $K$.

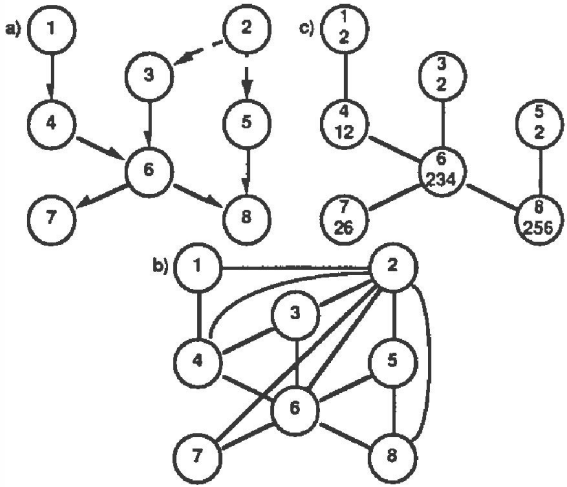

Figure 4: The Loop-Cutset Conditioning Method applied to the chest clinic example.

Another way to view Global Conditioning is that we are adding arcs in the moral graph from $K$ to all other nodes or, equivalently, that we are adding the conditioning set $K$ to every cluster in a cluster tree, since we must iterate over the possible values of $X_K$. Consider for example the belief network shown in Figure 4a, in which the arcs from 2 have been cut. Because we will be iteratively solving the new problem for every possible case of $X_2$, it is really as if an undirected arc had been added from 2 to all other nodes in the moral graph shown in Figure 4b, making it a chordal graph. A cluster tree for that graph is shown in Figure 4c. It could also have been obtained by building the cluster tree corresponding to the singly-connected network shown in Figure 4a (if we omit node 2) and then adding 2 to every cluster. This is not a particularly efficient cluster tree because only the (2, 6) arc was needed for triangulation. This cluster tree contains clusters with four elements, while the cluster trees for the same problem shown in Figure 2 have no clusters containing more than three elements. This result is formalized in the following theorem.

**Theorem 2 (Global Conditioning)**
*The Method of Global Conditioning is a special case of the Clustering Algorithm. It is less efficient than the Clustering Algorithm unless the conditioning set already appears in every cluster.*

**The Method of Loop-Cutset Conditioning** (Pearl 1986b; Peot and Shachter 1991) is closely related to the Method of Global Conditioning: the instantiated problems are solved using the (modified) Polytree Algorithm, so the conditioning set must render the original belief network at most singly connected. The set is called a **loop-cutset** because every undirected cycle (or **loop**) in the network contains at least one arc emanating from one of the nodes in the conditioning set (Pearl 1986a; Peot and Shachter 1991; Suermondt and Cooper 1988; Suermondt and Cooper 1990).

**Theorem 3 (Loop-Cutset Conditioning)** *The Method of Loop-Cutset Conditioning (Pearl 1986a) in modified form (Peot and Shachter 1991) is a special case of the Clustering Algorithm. It is less efficient than the Clustering Algorithm unless the conditioning set already appears in every family in the belief network.*

**Proof:** We know from Theorem 1 that the Polytree Algorithm is a special case of the Clustering Algorithm, with clusters corresponding to the families in the belief network, so the results follow from Theorem 2, after we construct the cluster tree. Start with whatever singly connected network will be used for the Polytree Algorithm. The cluster tree has the same topology, with the cluster corresponding to a node $j$ in the tree formed from the union of node $j$, its parents in the tree, and the conditioning set. This is a cluster tree because (1) it is a tree (or forest) by construction; (2) it is a join tree because the conditioning set $K$ is in every cluster and the cluster corresponding to every node $j \in K$ must be adjacent to the clusters of $j$'s children (every other cluster in which $j$ appears); and (3) every family is contained in some cluster, because the only arcs cut are from nodes in $K$ which are included in every cluster anyway. □

It might seem counterintuitive that Loop-Cutset Conditioning is related to the Clustering Algorithm. Consider the network shown in Figure 5a suggested by Pearl in the discussion following (Lauritzen and Spiegelhalter 1988). A "standard" chordal graph for this problem in shown in Figure 5b. Although conditioning on $X_1$ appears to be superior, it in fact corresponds exactly to the chordal graph shown in Figure 5c, which has the same number of cliques, but one with four elements instead of three. Pearl's suggestion that combining nodes, as shown in Figure 5d, might be superior, actually results in the chordal graph shown in Figure 5e, which is even less efficient than the other two chordal graphs.

## 5 PARALLEL IMPLEMENTATION OF GLOBAL CONDITIONING

In the preceding section it was shown that the Method of Global Conditioning is never more efficient than the Clustering Algorithm, but it can nonetheless present some natural opportunities for parallel implementation. In this section we show some of the technical details and some efficiencies possible in such an implementation.

We start with a **master cluster tree**, on which we would like to perform inference. From this master tree, we build **instantiated cluster trees** for each possible case $x_K$ of the conditioning variables $X_K$. After distributing evidence within the instantiated cluster trees, we construct a new master tree equivalent to what the original master tree would have been if we had performed message passing directly on it. Each instantiated cluster tree has the same topology as the master tree, but without the variables $X_K$. We fill in the table for each cluster (and separation set) in the instantiated tree by copying the table from the master tree corresponding to the particular values of $X_{S_i \cap K}$. Whenever there are some conditioning variables in a cluster in the master tree, $S_i \cap K \neq \emptyset$, the corresponding cluster in the instantiated tree should be marked as changed, since it now contains likelihood evidence about $X_K$. After passing messages within the instantiated tree, we can add the resulting tables into the master tree simply by summing out the variables $X_{K \setminus S_i}$ that do not appear in the master tree,

$$\Pr\{X_{S_i}, \epsilon\} = \sum_{K \setminus S_i} \Pr\{X_{S_i \cup K}, \epsilon\}.$$

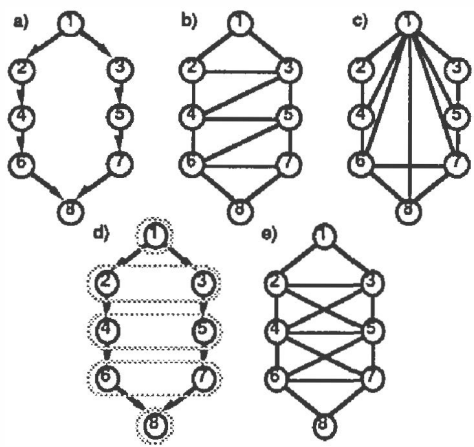

Figure 5: Pearl's Example for Clique Formation.

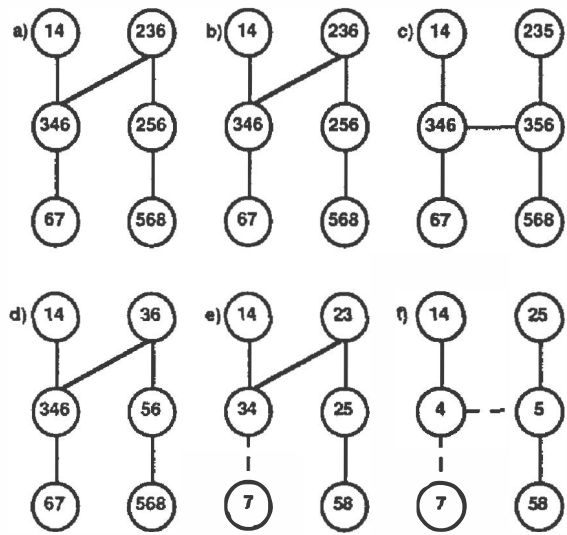

Figure 6: Master and Instantiated Cluster Trees.

The choice of master cluster tree depends on the conditioning set which will be used to construct the instantiated cluster trees. Consider the example shown in Figure 1a. If we are going to condition on node 2, we might want the master tree shown in Figure 6a rather than the one shown in Figure 6c or any of the ones shown in Figure 2. Although the tree shown in Figure 6a is a valid cluster tree based on the belief network, it could not be obtained directly from any of the cluster trees in Figure 2. Its corresponding instantiated cluster tree conditioned on node 2 is shown in Figure 6d. The instantiated cluster tree shown in Figure 6e corresponds to the master tree shown in Figure 6b conditioned on node 6. The instantiated tree shown in Figure 6f corresponds to the master tree shown in Figure 6c conditioned on nodes 3 and 6.

When we build an instantiated cluster tree, any separation sets which were subsets of the conditioning set become empty, as shown by the dashed arcs in Figure 6e and Figure 6f. This can be handled within the Clustering Algorithm as scalar separation sets, but there is some efficiency to be gained by exploiting the fact that the problem has been separated into "islands." Instead of passing the scalar messages throughout the network, where we will have to update the tables (in all but one island) twice, we can recognize the cumulative contributions among islands at a higher level and perform the scalar update only once, as we copy the table back into the master problem. To perform this, we need to construct a scalar "island tree" with a node for each island. We can compute the scalar value for any island by summing over its smallest cluster (or separation set). The messages sent between islands are then computed by the Clustering Algorithm, obtaining an update factor for each island, to be applied when the island's tables are copied back into the master cluster tree. If we ever encounter a zero scalar value for any island, then we can skip the rest of the processing for the entire instantiated cluster tree, because our model says that this particular instantiation is not possible given our observations $\epsilon$.

### Trading Time for Memory

Another application of the Method of Global Conditioning arises when there is insufficient memory to store the master cluster tree. The procedure from the previous section can be changed to trade extra compute time for memory savings. Thus, as in matrix algorithms, it is possible to transform a computation over a spatial dimension into an iteration of the values of that dimension over a spatially reduced problem.

If the master cluster tree cannot be explicitly constructed because of space limitations, we can modify the parallelized algorithm. First we must determine a good conditioning set and a good structure for the instantiated cluster tree, with cluster tables small enough for the restricted memory. We then serially consider each case of the conditioning variables. For each instantiation we initialize the cluster tree from the input distributions rather than from a master tree. If our goal is to compute $\Pr\{X_J, \epsilon\}$ for some set of variables $J$, then we accumulate the answer into these tables instead of into a master tree, but using the same kind of mapping.

## 6 RECOGNIZING NEW CLUSTER TREES

Although Global Conditioning is only performing the same operations as the Clustering Algorithm, it can provide additional insight about a problem. In this section, we explore a subtle trick that emerges from an ingenious application of Loop-Cutset Conditioning (Suermondt and others 1990) to a "star-shaped" diagnostic network constructed using similarity networks (Heckerman 1990a; Heckerman 1990b).

In that problem there is a single variable (disease) whose observation splits the belief network into many small disconnected pieces. By conditioning on that variable, each piece can be solved independently. (They are still dependent, but that dependence is reflected in their likelihood update for the disease node.) An abstract representation of the situation in cluster trees is shown in Figure 7, in which $X_0$ corresponds to the disease variable. Using the parallelization method in Section 5, we would be splitting this network into three "islands." The graphs shown in Figure 7a and Figure 7b are both valid cluster trees. If our goal is to collect at cluster 0 based on evidence recorded in clusters 067 and 12, the cluster tree shown in Figure 7b is superior to the one shown in Figure 7a. Of course, there is no way to anticipate exactly what evidence will be observed, so we can use the following result to restructure the cluster tree dynamically.

**Theorem 4 (Dynamic Restructuring)** *Given that there are two nonadjacent clusters $A$ and $B$ in a cluster tree, such that $A \cap B$ is equal to the separation set on some arc $e$ on the path between them, arc $e$ can be replaced by a new arc connecting $A$ and $B$. The change is completely reversible and messages for the old arc will be valid for the new one. If joint-form messages are used, then all messages in the network are still valid; otherwise, the messages along the old path from $A$ to $B$ might require updating.*

**Proof:** Because we have a join tree, $A \cap B$ must be contained in every cluster between $A$ and $B$. The arc $e$ can be replaced by an arc between any of the clusters along the path from $A$ until $e$ and any of the clusters along the path from $B$ until $e$ without violating the join tree property, and hence maintaining the cluster tree conditions. Because any pair of those clusters satisfy the same conditions as $A$ and B, the new arc can be drawn between $A$ and $B$ without loss of generality. The change is reversible, because the conditions will be satisfied by the two clusters which were incident to $e$ with respect to the new arc. The message in one direction along the old arc, with a slight abuse of notation, is

$$M_{AB} = \sum_{N \setminus B} \prod_{k \in K_{AB}} \Psi_k,$$

and it does not change because $K_{AB}$ is maintained.

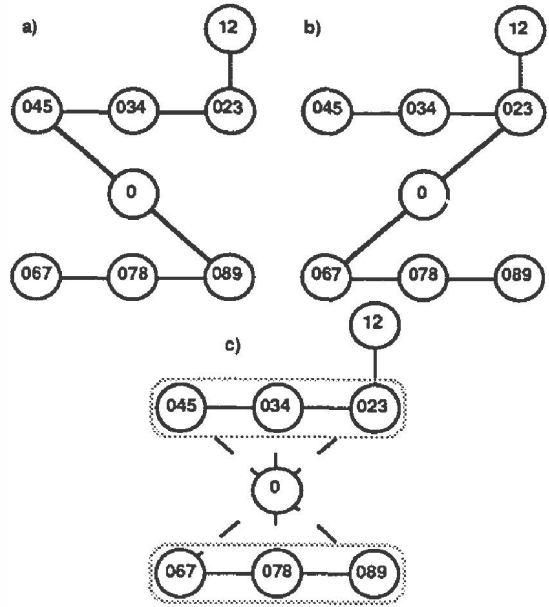

Figure 7: Dynamic Restructuring of Cluster Trees.

If joint-form messages are used, they can be interpreted as posterior joint distributions over the separation sets, and the separation sets are maintained. If factored form messages are used, their values depend on the $K_{ij}$ partition of the clusters in the network. The only place where these partitions change is along the old path between $A$ and $B$. □

This theorem can be easily applied. Consider the cluster tree shown in Figure 7c. Operationally, we can view the arc between 0 and the three clusters above it as "flexible," because any one of the three clusters can be connected to 0 without changing the messages on the arc. Likewise, there is a flexible arc between 0 and the three clusters below it. This is particularly powerful in combination with joint-form messages, although the updating formula for factored form messages is simple to compute because

$$P_{ij}^{\text{old}} = P_{ij}^{\text{new}}.$$

Not all of the new messages need to be computed if the only operation is to collect and not to distribute.

Another example of the theorem is in the cluster trees shown in Figure 2. In the tree shown in Figure 2b, the separation set between 67 and 346 is just 6, so the arc between them can be replaced by one from 67 to any other cluster containing 6. Thus we can immediately obtain the tree shown in Figure 2c, as well as one in which 67 is adjacent to 568 instead.

## 7  Conclusions

In this paper we have developed a generalization of Loop-Cutset Conditioning that we call Global Conditioning. We show that both are special cases of the Clustering Algorithm, a variation of the HUGIN algorithm (Jensen and others 1990a; Jensen and others 1990b). We use the insight from this relationship to develop a parallel and restricted memory version of the Clustering Algorithm. We also recognize an important class of problems in which we can simply and effectively restructure a cluster in response to a query. The Clustering Algorithm presented in this paper is not a new solution method, but rather a framework for comparison and substitution among methods. One is free to choose different clusters, different topologies for the cluster tree, different representations for messages, and different control strategies. We have much to learn about how to make and evaluate these choices.

Perhaps the most important choice from the list above is the selection of clusters, because it determines the efficiency of all of the methods. The key step is triangulation, in which the moral graph for the original belief network is made into a chordal graph. Although determining the "optimal" triangulation is NP-complete, there are excellent heuristics for this task (Kjærulff 1990). Some of the methods considered in this paper also present alternative triangulation strategies (Cooper 1990; Pearl 1986a; Pearl 1988; Suermondt and Cooper 1988; Suermondt and Cooper 1990).

One possible extension of these results would be to enhance the parallelized algorithm by recognizing when a particular instantiated cluster tree can be reused. This occurs when the variables in the cluster, $X_{S_i}$, are conditionally independent of some of the conditioning variables, $X_{K1}$, given the evidence $\epsilon$ and the remaining conditioning variables $X_{K2}$. (This independence can be recognized in linear time in the size of the network (Geiger and others 1990; Shachter 1988; Shachter 1990b).) If there is an efficient way to compute $\Pr\{X_{K1}|X_{K2},\epsilon\}$, then we can save iterations by recognizing that
$$\Pr\{X_{S_i \cup K},\epsilon\} = \Pr\{X_{K1}|X_{K2},\epsilon\}\Pr\{X_{S_i},X_{K2},\epsilon\}.$$

Finally, we have assumed throughout that the conditioning variables $X_K$ were taken through all their values. If, instead, an asymmetric tree of conditioning values is used, then we would be able to recognize some of the conditional independence in relations such as the "noisy or." This might lead to a global conditioning algorithm equivalent to Local Expressions (D'Ambrosio 1991).

### Acknowledgements

We benefited greatly from the comments of Derek Ayers, Richard Barlow, Greg Cooper, Bruce D'Ambrosio, Adnan Darwiche, Brendan Del Favero, Bob Fung, Perry McCarty, Mark Peot, Jaap Suermondt, and the anonymous referees.